\documentclass[runningheads]{llncs}
\usepackage[T1]{fontenc}
\usepackage{graphicx}
\usepackage{mathtools}
\usepackage{amsfonts}
\usepackage{xcolor}
\usepackage{hyperref}
\usepackage{booktabs}
\usepackage{array,multirow,graphicx}

\usepackage{color}

\begin{document}
%
\title{Enhancing Spatiotemporal Disease Progression Models via Latent Diffusion and Prior Knowledge}

%
%

\author{Lemuel Puglisi\inst{1} \and
Daniel C. Alexander\inst{2} \and
Daniele Ravì\inst{2,3}}
\authorrunning{L. Puglisi et al.}
\titlerunning{BrLP: Brain Latent Progression}
%
\institute{Dept. of Math and Computer Science, University of Catania, Italy \and
           Centre for Medical Image Computing, University College London, UK \and
           School of Physics, Engineering and CS, University of Hertfordshire, UK}

%

\maketitle              
\begin{abstract}
In this work, we introduce Brain Latent Progression (BrLP), a novel spatiotemporal disease progression model based on latent diffusion. BrLP is designed to predict the evolution of diseases at the individual level on 3D brain MRIs. Existing deep generative models developed for this task are primarily data-driven and face challenges in learning disease progressions. BrLP addresses these challenges by incorporating prior knowledge from disease models to enhance the accuracy of predictions. To implement this, we propose to integrate an auxiliary model that infers volumetric changes in various brain regions. Additionally, we introduce Latent Average Stabilization (LAS), a novel technique to improve spatiotemporal consistency of the predicted progression. BrLP is trained and evaluated on a large dataset comprising 11,730 T1-weighted brain MRIs from 2,805 subjects, collected from three publicly available, longitudinal Alzheimer's Disease (AD) studies. In our experiments, we compare the MRI scans generated by BrLP with the actual follow-up MRIs available from the subjects, in both cross-sectional and longitudinal settings. BrLP demonstrates significant improvements over existing methods, with an increase of 22\% in volumetric accuracy across AD-related brain regions and 43\% in image similarity to the ground-truth scans. The ability of BrLP to generate conditioned 3D scans at the subject level, along with the novelty of integrating prior knowledge to enhance accuracy, represents a significant advancement in disease progression modeling, opening new avenues for precision medicine. The code of BrLP is available at the following link: \url{https://github.com/LemuelPuglisi/BrLP}.

\end{abstract}

\section{Introduction}
Neurodegenerative diseases represent a global health challenge, affecting millions of people and leading to extensive morbidity and mortality. The situation is compounded by an increasingly ageing population, putting more strain on healthcare systems and society as a whole. Additionally, the progression of neurodegenerative diseases is characterized by its heterogeneous nature, with a variety of neuropathological patterns arising from different molecular subtypes~\cite{tijms2024cerebrospinal}. In particular, these diseases affect brain regions at varying rates and through distinct mechanisms, highlighting the intricate nature of their pathophysiology~\cite{young2018uncovering}. Therefore, we need to develop new methods aimed at better understanding disease development, which will pave the way for more targeted and personalized treatment strategies. Initial approaches used disease progression modeling primarily based on scalar biomarkers~\cite{young2024data,oxtoby2017imaging}. Despite the crude representations of these biomarkers, such approaches have been used to enhance our understanding of diseases~\cite{eshaghi2021identifying,vogel2021four}. A natural evolution of these efforts is developing spatiotemporal models, which represent disease progression using rich, high-dimensional imaging biomarkers operating directly on medical scans. Unlike scalar biomarkers, these solutions facilitate the visualization and precise localization of complex patterns of structural changes, thereby offering a more detailed understanding of disease dynamics. Recent approaches have leveraged deep generative techniques, such as Variational Autoencoders (VAEs)~\cite{sauty2022progression}, Generative Adversarial Networks (GANs)~\cite{ravi2022degenerative,xia2021learning,jung2021conditional,zhao2020prediction,pombo2023equitable}, and more recently, diffusion models~\cite{yoon2023sadm}, to infer the disease progression at the individual level. In particular, DaniNet~\cite{ravi2022degenerative} is a state-of-the-art model that uses adversarial learning combined with biological constraints to provide individualized predictions of brain MRIs. To mitigate memory requirements, DaniNet generates 2D slices that are then assembled into a 3D volume using a super-resolution module. Another approach is CounterSynth~\cite{pombo2023equitable}, a GAN-based counterfactual synthesis method that can simulate various conditions within a brain MRI, including ageing and disease progression. Lastly, SADM~\cite{yoon2023sadm} is a diffusion model designed to generate longitudinal scans through autoregressive sampling by using a sequence of prior MRI scans.  

\hypertarget{C1}{The} primary challenges of these methods are: 1) improving individualization by conditioning on subject-specific metadata; 2) using longitudinal scans when and if available; 3) enhancing spatiotemporal consistency to achieve a smooth progression across spatial and temporal dimensions; 4) managing the high memory demands imposed by the use of high-resolution 3D medical images~\cite{blumberg2018deeper}. Specifically, DaniNet~\cite{ravi2022degenerative} and CounterSynth~\cite{pombo2023equitable} are not able to directly use longitudinal data if accessible. SADM~\cite{yoon2023sadm} is not able to incorporate conditioning on subject-specific metadata and is also memory-intensive. Finally, neither CounterSynth nor SADM offer solutions to enforce spatiotemporal consistency. 

In response to these challenges, we introduce BrLP, a novel spatiotemporal model, offering several key contributions: i) we propose to combine an LDM~\cite{rombach2022high} and a ControlNet~\cite{zhang2023adding} to generate individualized brain MRIs conditioned on available subject data -- addressing challenge\hyperlink{C1}{ 1; }ii) we propose to integrate prior knowledge of disease progression by employing an auxiliary model designed to infer volumetric changes in different brain regions, allowing the use of longitudinal data when available -- addressing challenge\hyperlink{C1}{ 2; }iii) we propose LAS, a technique to improve spatiotemporal consistency in the predicted progression -- addressing challenge\hyperlink{C1}{ 3; }and iv) we use latent representations of brain MRIs to limit the memory demands for processing 3D scans -- addressing challenge\hyperlink{C1}{ 4.}

We evaluate BrLP by training it to learn progressive structural changes in the brains of individuals with different cognitive statuses: Cognitively Normal (CN), Mild Cognitive Impairment (MCI), and Alzheimer's Disease. To do so, we use a large dataset of 11,730 T1-weighted brain MRIs from 2,805 subjects, sourced from three publicly available longitudinal studies on AD. To the best of our knowledge, we are the first to propose a 3D conditional generative model for brain MRI that incorporates prior knowledge of disease progression into the image generation process.


\section{Methods}

\subsection{Background - Diffusion Models}
\label{sec:background} A Denoising Diffusion Probabilistic Model (DDPM)~\cite{ho2020denoising} is a deep generative model with two Markovian processes: forward diffusion and reverse diffusion. In the forward process, Gaussian noise is incrementally added to the original image $x_0$ over $T$ steps. At each step $t$, noise is introduced to the current image $x_{t-1}$ by sampling from a Gaussian transition probability defined as $q(x_t \mid x_{t-1}) \coloneqq \mathcal{N}(x_t; \sqrt{1 - \beta_t}x_{t-1}, \beta_t I)$, where $\beta_t$ follows a variance schedule. If $T$ is sufficiently large, $x_T$ will converge to pure Gaussian noise $x_T \sim \mathcal{N}(0, I)$. The reverse diffusion process aims to revert each diffusion step, allowing the generation of an image from the target distribution starting from pure noise $x_T$. The reverse transition probability has a Gaussian closed form, $q(x_{t-1} \mid x_t, x_0) = \mathcal{N}(x_{t-1} \mid \tilde\mu(x_0, x_t), \tilde\beta_t)$, conditioned on the real image $x_0$. As $x_0$ is not available during generation, a neural network is trained to approximate $\mu_\theta(x_t, t) \approx \tilde\mu(x_0, x_t)$. Following the work proposed in~\cite{ho2020denoising}, it is possible to reparameterise the mean in terms of $x_t$ and a noise term $\epsilon$, and then use a neural network to predict the noise $\epsilon_\theta(x_t, t) \approx \epsilon$, optimized with the following objective:

\begin{equation}
\label{eqn:ddpmloss}
\mathcal{L}_{\epsilon} \coloneqq \mathbb{E}_{t, x_t, \epsilon \sim \mathcal{N}(0, I)} \left[ 
\lVert \epsilon - \epsilon_\theta(x_t, t) \rVert^2 \right].
\end{equation}

An LDM~\cite{rombach2022high} extends the DDPM by applying the diffusion process to a latent representation $z$ of the image $x$, rather than to the image itself. This approach reduces the high memory demand while preserving the quality and flexibility of the models. The latent representation is obtained by training an autoencoder, composed of an encoder $\mathcal{E}$ and a decoder $\mathcal{D},$ such that the encoder maps the sample $x$ to the latent space $z = \mathcal{E}(x),$ and the decoder recovers it as $x = \mathcal{D}(z)$.

\begin{figure*}[t!]
    \centering
    \includegraphics[width=\textwidth]{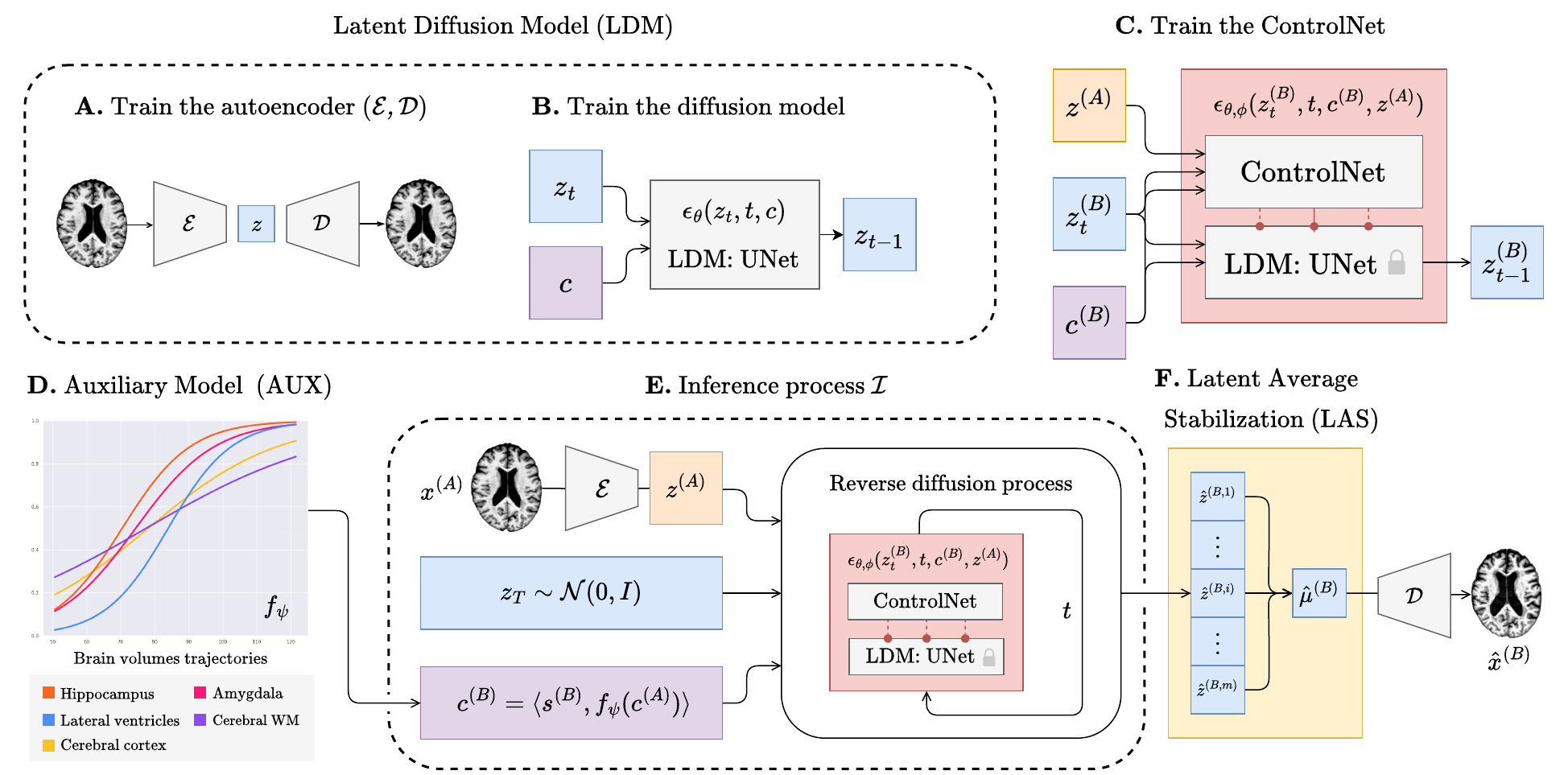}
    \caption{The overview of BrLP training and inference processes.}
    \label{fig:pipeline}
\end{figure*}

\subsection{Proposed Pipeline - Brain Latent Progression (BrLP)}
We now introduce the architecture of BrLP, comprising four key components: an LDM, a ControlNet, an auxiliary model, and a LAS block, each described in successive paragraphs. These four components, summarized in Figure~\ref{fig:pipeline}, collectively address the challenges outlined in the introduction. In particular, the LDM is designed to generate random 3D brain MRIs that conform to specific covariates, while ControlNet aims to specialize these MRI scans to specific anatomical structures of a subject. Additionally, the auxiliary model leverages prior knowledge of disease progression to improve the precision in predicting the volumetric changes of specific brain regions. Finally, the LAS block is used during inference to improve spatiotemporal consistency. Details concerning the training process and hyperparameter settings are provided in Table 1 of the Supplementary Material.

\subsubsection{LDM - Learning the brain MRIs distribution.}
Building upon~\cite{pinaya2022brain}, we train an LDM aimed to generate 3D brain MRIs mirroring specific covariates $c = \langle s, v \rangle$, where $s$ includes subject-specific metadata (age, sex, and cognitive status) while $v$ encompasses progression-related metrics such as volumes of brain regions (hippocampus, cerebral cortex, amygdala, cerebral white matter, and lateral ventricles) linked to AD progression~\cite{pini2016brain}. The construction of the LDM is a two-phase process. Initially, we train an autoencoder $(\mathcal{E}, \mathcal{D})$ (block A in Figure~\ref{fig:pipeline}) designed to produce a latent representation $z = \mathcal{E}(x)$ for each brain MRI $x$ within our dataset. Subsequently, we train a conditional UNet (block B in Figure~\ref{fig:pipeline}), represented as $\epsilon_\theta$, with network parameters $\theta$, aimed to estimate the noise $\epsilon_\theta(z_t, t, c)$ necessary for reverting from $z_t$ to $z_{t-1}$, as mentioned in Section \ref{sec:background}. We train $\epsilon_\theta$ by minimizing the loss $\mathcal{L}_{\epsilon}$ (Eq.~\ref{eqn:ddpmloss}). Covariates $c$ are integrated into the network as conditions using a cross-attention mechanism, in line with~\cite{rombach2022high}. The generation process initiates by sampling random Gaussian noise $z_T \sim \mathcal{N}(0,I)$ and then iteratively reverses each diffusion step $z_t \to z_{t-1}$ for $t=T, \dots, 1$. Decoding the output $z_0$ from the final step $t=1$ yields a synthetic brain MRI $\hat x = \mathcal{D}(z_0)$ that follows the specified covariates $c$.

\subsubsection{ControlNet - Conditioning on subject brain MRI.}
The LDM provides only a limited degree of control over the generated brain MRI via the covariates $c$, and it does not allow for conditioning the model on individual anatomical structures. The purpose of this block is to extend the capabilities of the LDM to encompass this additional control. To achieve this, we use ControlNet~\cite{zhang2023adding}, (block C in Figure~\ref{fig:pipeline}) a neural network designed to work in conjunction with the LDM. We conceptualize ControlNet and LDM as a unified network $\epsilon_{\theta,\phi}$, where $\theta$ represents the fixed network's parameters of the LDM and $\phi$ denotes the trainable network's parameters of ControlNet. As in the LDM, $\epsilon_{\theta,\phi}$ is still used to predict the noise $\epsilon_{\theta,\phi}(z_t, t, c, z)$ in the reverse diffusion step $z_t \to z_{t-1}$, now incorporating $z = \mathcal E(x)$ as a condition to encompass the structure of the target brain $x$ during the generation process. To train ControlNet, we use the latent representations $z^{(A)}$ and $z^{(B)}$ from pairs of brain MRIs of the same patient taken at different ages $A$ < $B$. The covariates $c^{(B)}$ associated with $z^{(B)}$ are known and used as target covariates. Each training iteration involves: i) sampling $t \sim U[1,T]$, ii) performing $t$ forward diffusion steps $z^{(B)} \to z_t^{(B)}$, iii) predicting the noise $\epsilon_{\theta,\phi}(z_t^{(B)}, t, c^{(B)}, z^{(A)})$ to revert $z^{(B)}_t \to z^{(B)}_{t-1}$, and iv) minimizing the loss $\mathcal{L}_{\epsilon}$ (Eq.~\ref{eqn:ddpmloss}).

\subsubsection{Proposed auxiliary model - Leveraging disease prior knowledge.}
\label{sec:auxiliarymodel}
AD-related regions shrink or expand over time and at different rates~\cite{pini2016brain}. Deep-learning-based spatiotemporal models strive to learn these progression rates directly from brain MRIs in a black-box manner, which can be very challenging. To aid this process, we propose incorporating prior knowledge of volumetric changes directly into our pipeline. To do so, we exploit an auxiliary model $f_\psi$ (block D in Figure~\ref{fig:pipeline}) able to predict how the volumes of AD-related regions change over time and provide this information to the LDM via the progression-related covariates $v$. The choice of our auxiliary model is tailored to two scenarios, making BrLP flexible for both cross-sectional and longitudinal data. For subjects with a single scan available at age $A$, we employ a regression model to estimate volumetric changes $\hat{v}^{(B)} = f_\psi(c^{(A)})$ at age $B$. For subjects with $n$ past visits accessible at ages $A_1, \dots, A_n$, we predict $\hat{v}^{(B)} = f\psi(c^{(A_1)}, \dots, c^{(A_n)})$ using Disease Course Mapping (DCM)~\cite{schiratti2017bayesian,koval2021ad}, a model specifically designed for disease progression. DCM is intended to provide a more accurate trajectory in alignment with the subject's history of volumetric changes available. While we employ DCM as a potential solution, any suitable disease progression model can be used in BrLP.

\subsubsection{Inference process.}
Let $x^{(A)}$ be the input brain MRI from a subject at age $A$, with known subject-specific metadata $s^{(A)}$ and progression-related volumes $v^{(A)}$ measured from $x^{(A)}$. As summarized in block E from Figure~\ref{fig:pipeline}, to infer the brain MRI $x^{(B)}$ at age $B > A$, we perform six steps: i) predict the progression-related volumes $\hat v^{(B)} = f_\psi(c^{(A)})$ using the auxiliary model; ii) concatenate this information with the subject-specific metadata $s^{(B)}$ to form the target covariates $c^{(B)} = \langle s^{(B)}, \hat v^{(B)}\rangle$; iii) compute the latent $z^{(A)} = \mathcal{E}(x^{(A)})$; iv) sample random Gaussian noise $z_T \sim \mathcal{N}(0,I)$; v) run the reverse diffusion process by predicting the noise $\epsilon_{\theta, \phi}(z_t, t, c^{(B)}, z^{(A)})$ to reverse each diffusion step for $t=T,\dots,1$; and finally vi) employ the decoder $\mathcal{D}$ to reconstruct the predicted brain MRI $\hat{x}^{(B)} = \mathcal{D}(z_0)$ in the imaging domain. This inference process is summarized into a compact notation $\hat z^{(B)} = \mathcal{I}(z_T, x^{(A)}, c^{(A)})$ and $\hat x^{(B)} = \mathcal{D}(\hat z^{(B)})$.

\subsubsection{Enhance inference via proposed Latent Average Stabilization (LAS).}
Variations in the initial value $x_T \sim \mathcal{N}(0,I)$ can lead to slight discrepancies in the results produced by the inference process. These discrepancies are especially noticeable when making predictions over successive timesteps, manifesting as irregular patterns or non-smooth transitions of progression. Therefore, we introduce LAS (block F in Figure~\ref{fig:pipeline}), a technique to improve spatiotemporal consistency by averaging different results of the inference process. In particular, LAS is based on the assumption that the predictions $\hat z^{(B)} = \mathcal{I}(z_T, x^{(A)}, c^{(A)})$ deviate from a theoretical mean $\mu^{(B)} = \mathbb{E}[\hat z^{(B)}]$. To estimate the expected value $\mu^{(B)}$, we propose to repeat the inference process $m$ times and average the results:
\begin{equation}
\mu^{(B)} =  \mathop{\mathbb{E}}_{z_T \sim \mathcal{N}(0,I)} \bigg[ \mathcal{I}(z_T, x^{(A)}, c^{(A)})\bigg] \approx \frac{1}{m}\sum^m \mathcal{I}(z_T, x^{(A)}, c^{(A)}).
\end{equation}
Similar to before, we decode the predicted scan as $\hat x^{(B)} = \mathcal{D}(\mu^{(B)})$. The entire inference process (with $m=4$) requires $\sim$4.8s per MRI on a consumer GPU.


\section{Experiments and Results}

\subsubsection{Data.} 
We collect a large dataset comprising 11,730 T1-weighted brain MRI scans from 2,805 subjects across various publicly available longitudinal studies: ADNI 1/2/3/GO (1,990 subjects)~\cite{petersen2010alzheimer}, OASIS-3 (573 subjects)~\cite{lamontagne2019oasis}, and AIBL (242 subjects)~\cite{ellis2009australian}. Each subject has at least two MRIs, and each scan is acquired during a different visit. Age, sex, and cognitive status were available from all datasets. The average age is $74 \pm 7$ years, and 53\% of the subjects are male. Based on the final visit, 43.8\% of subjects are classified as CN, 25.7\% exhibit or develop MCI, and 30.5\% exhibit or develop AD. We randomly split data into a training set (80\%), a validation set (5\%), and a testing set (15\%) with no overlapping subjects. The validation set is used for early stopping during training. Each brain MRI is pre-processed using: N4 bias-field correction~\cite{tustison2010n4itk}, skull stripping~\cite{hoopes2022synthstrip}, affine registration to the MNI space, intensity normalization~\cite{shinohara2014statistical} and resampling to 1.5 mm\textsuperscript{3}. The volumes used as progression-related covariates and for our subsequent evaluation are calculated using SynthSeg 2.0~\cite{billot2023synthseg} and are expressed as percentages of the total brain volume to account for individual differences.

\subsubsection{Evaluation metrics.}
We evaluate BrLP using image-based and volumetric metrics to compare the predicted brain MRI scans with the subjects' actual follow-up scans. In particular, the Mean Squared Error (MSE) and the Structural Similarity Index (SSIM) are used to assess image similarity between the scans. Instead, volumetric metrics in AD-related regions (hippocampus, amygdala, lateral ventricles, cerebrospinal fluid (CSF), and thalamus) evaluate the model's accuracy in tracking disease progression. Specifically, the Mean Absolute Error (MAE) between the volumes of actual follow-up scans and the generated brain MRIs is reported in the results. Notably, CSF and thalamus are excluded from progression-related covariates, enabling the analysis of unconditioned regions in our predictions.

\subsubsection{Ablation study.}
We conduct an ablation study to assess the contributions of: i) the auxiliary model (AUX) and ii) the proposed technique for spatiotemporal consistency (LAS). The results are presented at the top of Table~\ref{tab:performance}. BrLP without AUX and LAS is referred to as ``base''. The experiments demonstrate that both LAS and AUX enhance performance, reducing volumetric errors by 5\% and 4\%, respectively. An example of improvement achieved with LAS is provided in Figure 2 of the Supplementary Material. Employing both AUX and LAS together offers the optimal setup, achieving an average reduction in volumetric error of 7\%. This optimal configuration is used for comparisons against other approaches, with the only variation being the type of auxiliary model used in our pipeline.

\begin{table}[t]
    \caption{Results from the ablation study and comparison with baseline methods. MAE (± SD) in predicted volumes is expressed as a percentage of total brain volume.}
    \label{tab:performance}
    \setlength{\tabcolsep}{5pt}
    \def\arraystretch{1.5}
    \resizebox{\columnwidth}{!}{
    
    \begin{tabular}{c|l|c|cc|ccc|cc} \hline 
    &&Config. & \multicolumn{2}{c|}{\textbf{Image-based metrics}} & \multicolumn{3}{c|}{\textbf{MAE (conditional region volumes)}} & \multicolumn{2}{c}{\textbf{MAE (unconditional reg. volumes)}} \\
    &\textbf{Method}& (AUX) & MSE $\downarrow$ & SSIM $\uparrow$ & Hippocampus $\downarrow$ & Amygdala $\downarrow$ & Lat. Ventricle $\downarrow$ & Thalamus $\downarrow$ & CSF $\downarrow$ \\
    
    \hline
    \multirow{4}{*}{\rotatebox[origin=c]{90}{\parbox[c]{2.3cm}{\centering \textbf{Ablation}}}}&Base & - & 0.005 ± 0.003 & 0.89 ± 0.03 & 0.026 ± 0.023 & 0.016 ± 0.015 &0.279 ± 0.347 & 0.030 ± 0.023 & 0.889 ± 0.681 \\
    &Base + AUX & LM & 0.005 ± 0.002 & 0.90 ± 0.03 & 0.024 ± 0.021 & 0.015 ± 0.013 & 0.279 ± 0.314 & 0.030 ± 0.024	& 0.851 ± 0.632 \\
    &Base + LAS & - & 0.005 ± 0.002 & 0.90 ± 0.03 & 0.025 ± 0.022 & 0.015 ± 0.014 & 0.258 ± 0.330 & \textbf{0.029 ± 0.022} & 0.851 ± 0.659\\
    &Base + LAS + AUX & LM & \textbf{0.004 ± 0.002} & \textbf{0.91 ± 0.03} & \textbf{0.023 ± 0.021} & \textbf{0.015 ± 0.014} & \textbf{0.255 ± 0.303} & \textbf{0.029 ± 0.023} & \textbf{0.829 ± 0.624}\\
    \specialrule{2pt}{1pt}{1pt}
    \multirow{6}{*}{\rotatebox[origin=c]{90}{\parbox[c]{4cm}{\centering \textbf{Comparison Study}}}}
    & \multicolumn{9}{l}{\textbf{Single-image (Cross-sectional)}} \\
    \cline{2-10}
    &DaniNet~\cite{ravi2022degenerative} & - & 0.016 ± 0.007 & 0.62 ± 0.16 & 0.030 ± 0.030 & 0.018 ± 0.017 & 0.257 ± 0.222 & 0.038  ± 0.030 & 1.081 ±  0.814 \\
    &CounterSynth~\cite{pombo2023equitable} & - & 0.010 ± 0.004 & 0.82 ± 0.05 & 0.030 ± 0.018 & \textbf{0.014 ± 0.010} & 0.310 ± 0.311 & 0.127 ± 0.035&0.881 ± 0.672 \\
    &BrLP (Proposed) & LM & \textbf{0.004 ± 0.002} & \textbf{0.91 ± 0.03} & \textbf{0.023 ± 0.021} & 0.015 ± 0.014 & \textbf{0.255 ± 0.303} & \textbf{0.029 ± 0.023} & \textbf{0.829 ± 0.624}\\
    
    \cline{2-10}
    & \multicolumn{9}{l}{\textbf{Sequence-aware (Longitudinal)}} \\
    \cline{2-10}
    &Latent-SADM~\cite{yoon2023sadm} & - & 0.008 ± 0.002 & 0.85 ± 0.02 & 0.035 ± 0.027 & 0.018 ± 0.015 & 0.329 ± 0.328 & 0.037 ± 0.028 & 0.924 ± 0.705 \\
    &BrLP (Proposed) & DCM & \textbf{0.004 ± 0.002} & \textbf{0.91 ± 0.03} & \textbf{0.020 ± 0.017} & \textbf{0.014 ± 0.013} & \textbf{0.240 ± 0.259} & \textbf{0.031 ± 0.024} & \textbf{0.810 ± 0.631}\\
    \end{tabular}
    }
\end{table}

\begin{figure*}[h!]
    \centering
    \includegraphics[width=0.95\linewidth]{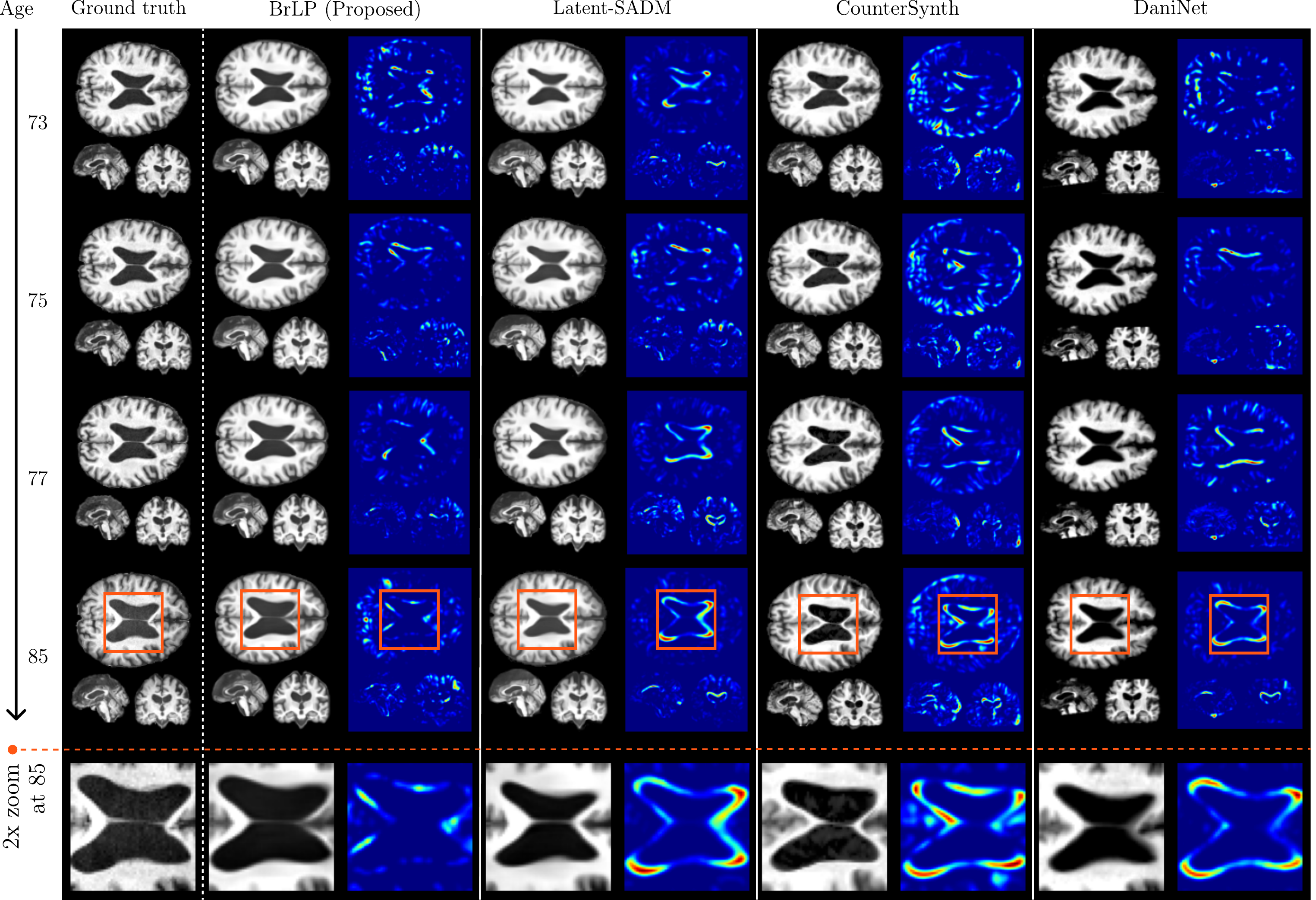}
    \caption{A comparison between the real progression of a 70 y.o. subject with MCI over 15 years and the predictions obtained by BrLP and the baseline methods. Each method shows a predicted MRI (left) and its deviation from the subject's real brain MRI (right).}
    \label{fig:comparison}
\end{figure*}

\subsubsection{Comparison with baselines.} \label{sec:baseline}
We categorize existing methods into single-image (cross-sectional) and sequence-aware (longitudinal) approaches. Single-image approaches, such as DaniNet~\cite{ravi2022degenerative} and CounterSynth~\cite{pombo2023equitable}, predict progression using just one brain MRI as input. Sequence-aware methods, like SADM~\cite{yoon2023sadm}, leverage a series of prior brain MRIs as input. Due to the large memory demands of SADM, we have re-implemented it using an LDM, allowing the comparisons in our experiments. We refer to it as Latent-SADM. To evaluate all these methods, we conduct two separate experiments. In single-image methods, we predict all subsequent MRIs for a subject based on their initial scan. For sequence-aware methods, we use the first half of a subject's MRI visits to predict all subsequent MRIs in the latter half. In single-image settings, our approach uses a Linear Model (LM) as the auxiliary model. In contrast, for sequence-aware experiments, we employ the last available MRI in the sequence as the input for BrLP and fit a logistic DCM on the first half of the subject's visits as the auxiliary model.

Results from our experiments are presented in Table 1. We observe an average decrease of 62\% (SD = 10\%) in MSE and an average increase of 43\% (SD = 18\%) in SSIM compared to other baselines. In terms of volumetric measurements across various brain regions, our method shows improvements of 17.55\% (SD = 8.79\%) over DaniNet, 23.40\% (SD = 28.85\%) over CounterSynth, and 24.14\% (SD = 10.63\%) over Latent-SADM. We did not observe any particular differences in the improvement obtained in conditioned and non-conditioned regions. Additionally, paired t-tests (\textit{p} < 0.001) verified the statistical significance of the observed improvements. Finally, Figure~\ref{fig:comparison} presents a visual comparison between the actual progression of a 70-year-old subject over 15 years and the predictions obtained by BrLP and the baseline methods. The results from Latent-SADM and DaniNet exhibit a spatiotemporal mismatch in predicting the lateral ventricles' enlargement, whereas CounterSynth fails to capture the structural changes observed in the real progression. On the other hand, BrLP shows the most accurate prediction of the brain's anatomical changes, confirming the previous quantitative findings. It is worth noting that we observed limitations in LAS performance in underrepresented conditions, such as ages over 90 years, resulting in slightly non-monotonic progression (see Case Study 4 in our supplementary video).


\section{Conclusion}
In this work, we propose BrLP, a 3D spatiotemporal model that accurately captures the progression patterns of neurodegenerative diseases by forecasting the evolution of 3D brain MRIs at the individual level. While we have showcased the application of our pipeline on brain MRIs, BrLP holds potential for use with other imaging modalities and to model different progressive diseases. Importantly, our framework can be easily extended to integrate additional covariates, such as genetic data, providing further personalized insights into our predictions.


%
%
%
\bibliographystyle{splncs04}
\bibliography{bibliography}
\end{document}